\documentclass{llncs}

\usepackage{times}
\usepackage{epsfig}
\usepackage{graphicx}
\usepackage{amsmath}
\usepackage{amssymb}
\usepackage{subcaption}
\usepackage{comment}
\usepackage{xcolor}

\graphicspath{{./images/}}

\usepackage[pagebackref=true,breaklinks=true,letterpaper=true,colorlinks,bookmarks=false]{hyperref}

\usepackage{xspace}

\makeatletter
\DeclareRobustCommand\onedot{\futurelet\@let@token\@onedot}
\def\@onedot{\ifx\@let@token.\else.\null\fi\xspace}

\def\eg{\emph{e.g}\onedot} 
\def\ie{\emph{i.e}\onedot}

\def\etal{\emph{et al}\onedot}
\makeatother

\begin{document}

\title{Indirect Match Highlights Detection with Deep Convolutional Neural Networks}

\author{Marco Godi\inst{1} \and Paolo Rota\inst{2} \and Francesco Setti\inst{1,3}}

\institute{
Department of Computer Science, University of Verona, Verona, Italy
\and Pattern Analysis and Computer Vision (PAVIS), Italian Institute of Technology, Genova, Italy
\and Institute of Cognitive Science and Technology, National Research Counsil, Trento, Italy\\
\email{francesco.setti@univr.it}
}

\authorrunning{M.~Godi \emph{et al.}}

\maketitle

\begin{abstract}
  Highlights in a sport video are usually referred as actions that stimulate excitement or attract attention of the audience. A big effort is spent in designing techniques which find automatically  highlights, in order to automatize the otherwise manual editing process.
  Most of the state-of-the-art approaches try to solve the problem by training a classifier using the information extracted on the tv-like framing of players playing on the game pitch, learning to detect game actions which are labeled by human observers according to their perception of highlight. Obviously, this is a long and expensive work. In this paper, we reverse the paradigm: instead of looking at the gameplay, inferring what could be exciting for the audience, we directly analyze the audience behavior, which we assume is triggered by events happening during the game.
  We apply deep 3D Convolutional Neural Network (3D-CNN) to extract visual features from cropped video recordings of the supporters that are attending the event. Outputs of the crops belonging to the same frame are then accumulated to produce a value indicating the Highlight Likelihood (HL) which is then used to discriminate between positive (\ie when a highlight occurs) and negative samples (\ie standard play or time-outs).
  Experimental results on a public dataset of ice-hockey matches demonstrate the effectiveness of our method and promote further research in this new exciting direction.
\end{abstract}

\section{Introduction}
\label{sec:intro}

Sport video summarization, or highlights generation, is the process of creating a synopsis of a video of a given sport event that gives the viewer a general overview of the whole match. This process incorporates two different tasks: (1) to detect the most important moments of the event, and (2) organize the extracted content into a limited display time.
While the second point is a widely-known problem in the multimedia and broadcasting community, the definition of \emph{what is a highlight} has different interpretations in the community.
According to~\cite{hanjalic2005tmm}, highlights are ``those video segments
that are expected to excite the users the most''. In~\cite{zhu2007tmm}, the focus relaxes from excitement to general attention, and thus salient moments are the ones that attract audience attention the most. These two definitions would imply to explicitly design specific models for extracting excitment from the crowd in one case and attention on the other.  In this paper we overcome this problem by automatically learn visual features using deep architectures that discriminate between highlights and ordinary actions.


\begin{figure}[t]
	\centering
	\begin{subfigure}[b]{.45\columnwidth}
  		\centering
  		\includegraphics[width=\linewidth]{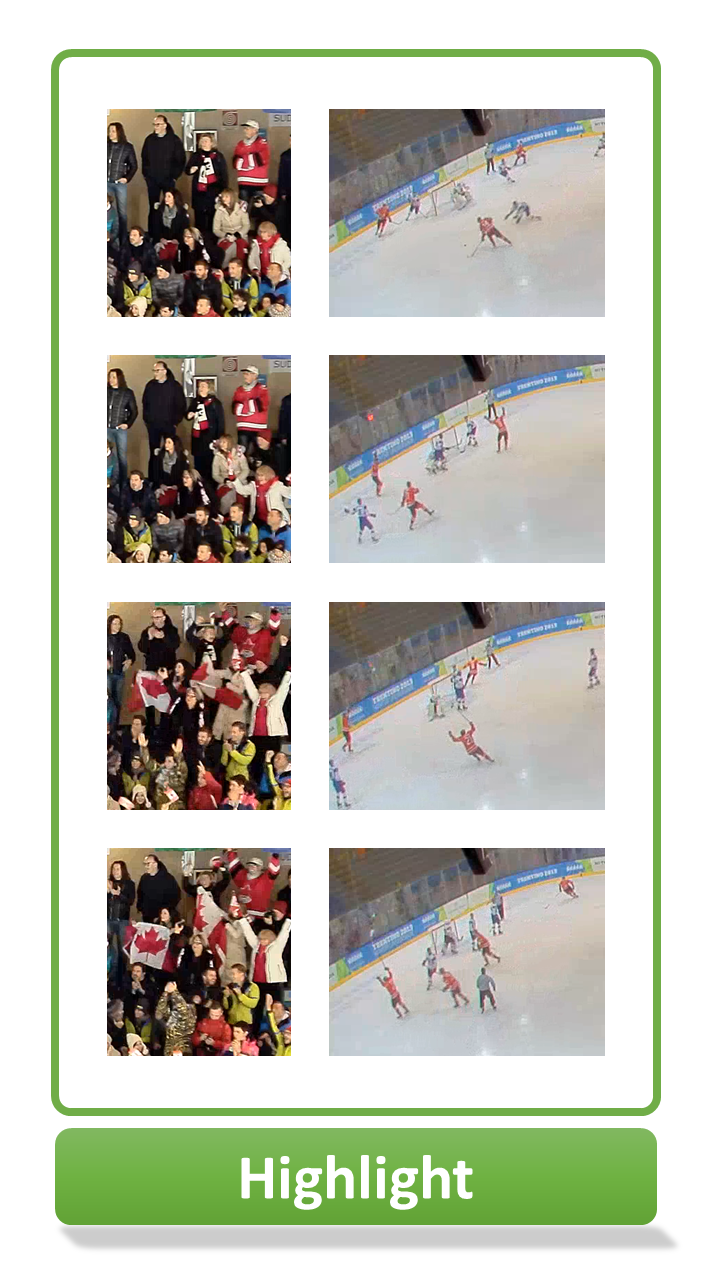}
	\end{subfigure}
	\begin{subfigure}[b]{.45\columnwidth}
  		\centering
  		\includegraphics[width=\linewidth]{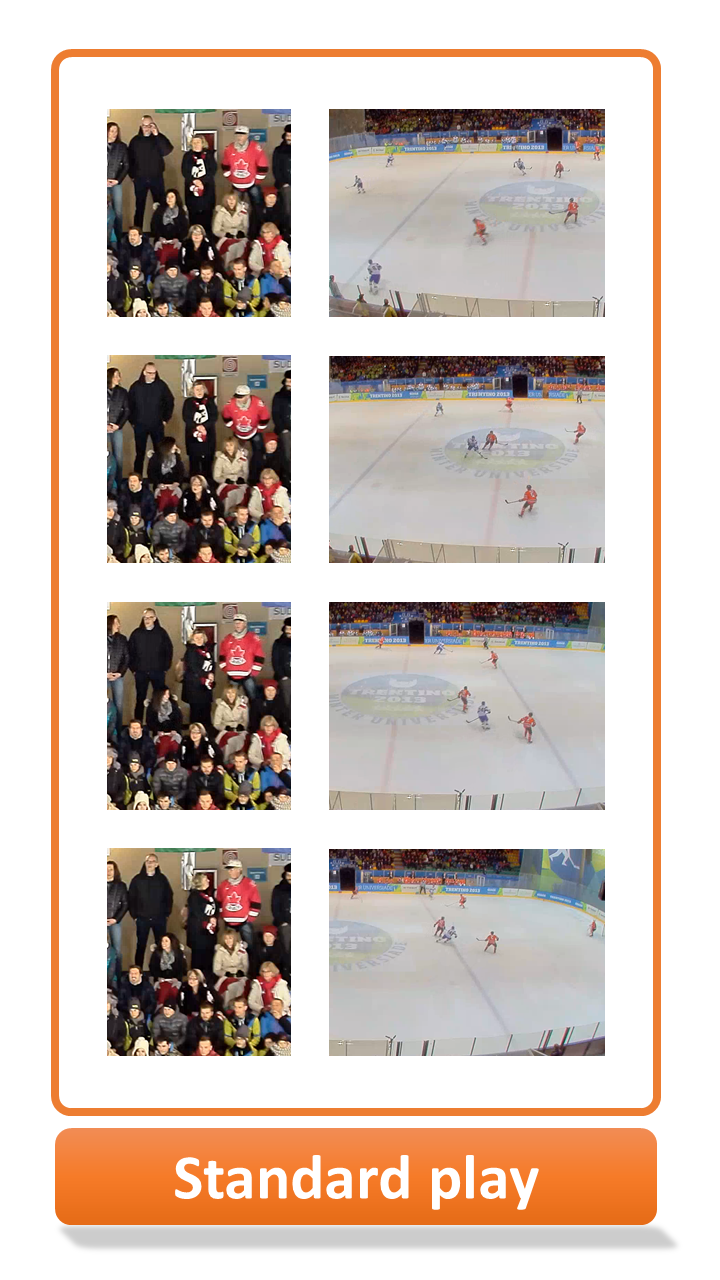}
	\end{subfigure}
	\caption{Example video sequences of a goal event (left) and standard play time (right).}
	\label{fig:excrop}
\end{figure}

Traditionally, extracting sport highlights has been a labor intensive activity, primarily because it requires good judgment to select and define salient moments throughout the whole game.
Then, highlights are manually edited by experts, to generate a video summary that is significant, coherent and understandable by humans.
State-of-the-art artificial intelligence is still far away from having solved the whole problem.

In the last years, there has been an increasing demand for automatic and semi-automatic tools for highlights generation, mainly due to the huge amount of data (\ie sport event videos) generated every day and made available through the Internet.
Specialized broadcasters and websites are able to deliver sport highlights minutes after the end of the event, handling thousands of events every day.
As a consequence, there has been extensive research in this area, with the development of several techniques based on image and video processing~\cite{bertini2003mir,chauhan2016ngct,hanjalic2003icip,hanjalic2005tmm,nguyen2014mmsp,tjondronegoro2004mm,zhu2007tmm}.
More recently, many works started using additional sources of information to increase performances, including audio recordings~\cite{rui2000acmmm,xiong2003icassp}, textual narratives~\cite{suksai2016icsec}, social networks~\cite{fiao2016ace,hsieh2012icme,tang2012chi}, and audience behavior~\cite{conigliaro2013attento,conigliaro2013viewing,conigliaro2013observing,peng2011tmm}.
Despite some solutions are already present on the market, performances are in general still fairly poor and we believe there is room for new research on this topic.\\

While previous work attempted to detect in sport videos actions that stimulate excitement~\cite{hanjalic2003icip} or attract attention~\cite{zhu2007tmm} of the audience, in this paper we reverse the problem by analyzing the audience behavior to identify changes in emotions, that can only be triggered by highlights on the game field.

Specifically, we present a novel approach for sport highlight generation which is based on the observation of the audience behavior. This approach is based on the analysis of a set of space-time cuboids using a 3D-CNN architecture. All the samples are trained singularly, the result for each cuboid at a certain time step is then processed through an accumulator which generates a sort of highlight probability for the whole audience that will be used to perform the final ranking.

The rest of the paper is organized as follows: in Section~\ref{sec:soa} we briefly present the state-of-the-art in automatic highlight detection. In Section~\ref{sec:method} we detail the proposed methodology, while in Section~\ref{sec:exp} we show some qualitative and quantitative results on a public dataset of hockey matches. Lastly, in Section~\ref{sec:concl} we draw some conclusions and perspectives for future works.

\section{Related work}
\label{sec:soa}

Money and Angius~\cite{money2008vcir} provide an extensive literature survey on video summarization. According to the taxonomy proposend in that paper, related work can be classified into three categories: (1) internal summarization techniques; 2) external summarization techniques; and 3) hybrid summarization techniques.
By definition, \emph{internal summarization techniques} rely only on information provided by the video (and audio) streams of the event. These techniques extract low-level image, audio, and text features to facilitate summarization and for several years have been the most common summarization techniques.
\emph{External summarization techniques} require additional sources of information, not contained in the video streams. These are usually user-based information --\ie information provided directly from users-- and contextual information --such as the time and location in which the video was recorded.
As for \emph{hybrid summarization techniques}, both internal and external information are
analyzed, allowing to reduce the semantic gap between the low level features and the semantic concepts.

\textbf{Social networks.}
According to Hsieh \etal~\cite{hsieh2012icme}, the quantity of comments and re-tweets can represent the most exciting moments in a sport event. A highlight can be determined by analyzing the keywords in the comments and observing if the number of comments and re-tweets passes a certain threshold.
Fi\~{a}o \etal~\cite{fiao2016ace} uses emotions shared by the spectators during the match via social networks to build a system capable of generating automatic highlight videos of sports match TV broadcasts. Auxiliary sources of information are TV broadcast videos, the audio, the analysis of the movement and manual annotations (when available). The system also allows for the user to query the video to extract specific clips (\eg attacking plays of a specific team).

\textbf{Text.}
In~\cite{suksai2016icsec}, Suksai and Ratanaworabhan propose an approach that combines on-line information retrieval with text extraction using OCR techniques. This way, they are able to limit the number of false positives.

\textbf{Audio.}
Rui \etal~\cite{rui2000acmmm} presents a method that uses audio signals to build video highlights for baseball games. It analyzes the speech of the match announcer, both audio amplitude and voice tone, to estimate whether the announcer is being excited or not. In addition, the ambient sound from the surrounding environment and the audience are also taken into considerations.
Built on this work, Xiong \etal~\cite{xiong2003icassp} handpicked the highlight events and analyzed the environment and audience sounds at each of those highlight events. They discovered that there exists a strong correlation between loud and buzzing noise and some major highlight events. This correlation exists in all the three sports being analyzed: baseball, golf, and soccer.

\textbf{Audience.}
Peng \etal~\cite{peng2011tmm} propose the Interest Meter (IM), a system able to measure user’s interest and thus use it to conduct video summarization. The IM takes account attention states (\eg eye movement, blink, and head motion) and emotion states (\eg facial expression). These features are then fused together by a fuzzy fusion scheme that outputs a quantitative interest score, determine interesting parts of videos, and finally concatenate them as video summaries.
In~\cite{conigliaro2013attento}, Conigliaro \etal use motion cues (\ie optical flow intensity and direction entropy) to estimate the excitement level of audience of a team sport event and to identify groups of supporters of different teams. In~\cite{conigliaro2013viewing}, these features are used to identify highlights in team sport events using mean shift clustering.

\begin{figure*}[t!]
 \centering
 \includegraphics[width=\textwidth]{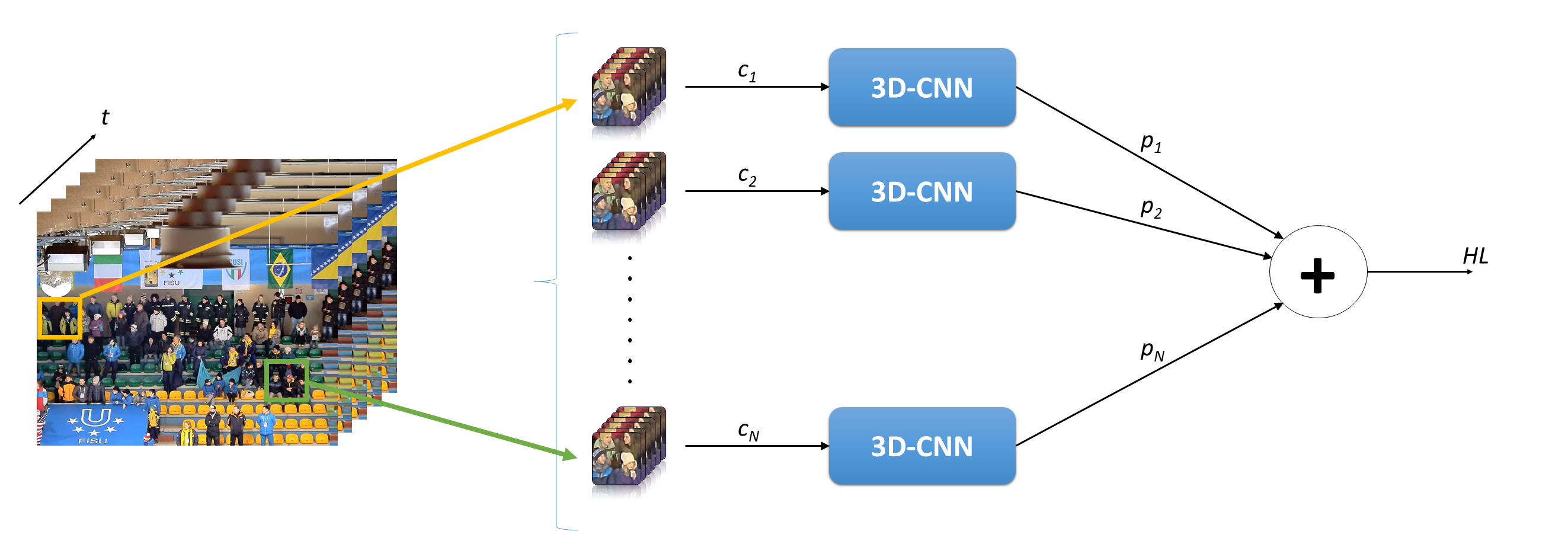}
 \caption{Sketch of the overall method.}
 \label{fig:cnn}
\end{figure*}

\section{Method}
\label{sec:method}

The proposed highlights detection methodology uses a 3D Convolutional Neural Network (3D-CNN) to extract visual features from video recordings of the audience of the event, and classify them in positive samples (\ie when a highlight occurs) and negative samples (\ie standard play or timeouts).

From empirical observations, the audience reaction of a highlight (\eg a goal) lasts for at least the 10 seconds that follows the event itself. For this reason, temporal resolution is not a critical parameter and downsampling the video from 30 to 3 fps allowed us to reduce the computational burden without losing the informative part of the video. The 3D-CNN cuboid is extracted from a manually selected rectangular area that roughly contained the bulk of the audience, using a uniform grid with fixed spatial dimension of 100$\times$100 pixels, while the temporal resolution has been set to 30 frames. These parameters are the result of an a priori intuition that each block should be able to represent a portion of spectators which should not be too large, in order to reduce the computational burden,  but at the same time it should not be too small since this would bring to be too much location dependent.
For our model we used a sliding window with a stride of 50 pixels resulting in a maximum overlap between two crops of 50\%

In order to detect and rank the most important moments in the video sequence we follow the idea of Conigliaro et al. \cite{conigliaro2015cvpr}, where information accumulators along time have been proposed to segment supporters of the two different playing teams. Our goal is however different: unlike them, we are interested in a global analysis of the excitement of the audience regardless of the supporting preference at a certain time. For this reason we are using an accumulator strategy over the whole audience location in the scene.
Each spatio-temporal cuboid $C_i$, $i=1,...,N$ represents a sample that is fed into a 3D-CNN and analyzed independently; then, for each time instant the related probability score $p_i$, $i=1,...,N$ of being a positive class is accumulated over all the samples in the spatial dimension, generating a scalar value representing the \emph{Highlight Likelihood} (HL) that is a score representing how likely a particular instant can be considered an highlight or not. A sketch of the overall system is shown in Fig.~\ref{fig:cnn}. 


\subsection{Network Architecture}

Inspired by earlier works on action recognition~\cite{ji2013pami,tran2015iccv}, we use a 3D Convolutional Neural Network composed by 4 convolutional and 3 fully connected layers.

The network takes as input video cuboids of 100$\times$100$\times$30, where the first two numbers refer to the spatial dimension while the third is the temporal depth (number of frames).
The first two convolutional layers are composed 12 filters 3$\times$3$\times$3, to capture spatio-temporal features from the raw data. These are followed by a 2$\times$2$\times$2 max pooling layer to detect features at different scales.   
In the latter two convolutional layers, 8 3$\times$3$\times$3 convolutional filters have been used. In all convolutional layers the ReLU activation has been used. The network is then unfolded with a flatten layer followed by three fully connected layers of decreasing dimensionality (32, 8, and 2 neurons respectively). The final classification task is achieved by a softmax layer that outputs the probability of a test sample to belong to each of the two classes: ``highlight'' and ``standard play''.


\section{Experiments}
\label{sec:exp}

In this section we provide both qualitative and quantitative results to validate our proposed methodology. For the evaluation we adopted the S-Hock dataset~\cite{setti2017cviu}, a publicly available dataset composed by 6 ice-hockey games recorded during the Winter Universiade held in Trentino (Italy) in 2013. This dataset, besides a set of short videos heavily annotated on low level features (\eg people bounding boxes, head pose, and action labels), it provides also a set of synchronized multi-view full matches with high-level event annotation. In these games, the labeling consist in the time position of meaningful events such as goals, fouls, shots, saves, fights and timeouts.

In this work we considered only two matches: the final match (Canada-Kazakhstan) which is used for training the neural network, and the semi-final match (USA-Kazakhstan), used for testing.

\subsection{3D-CNN training procedure}

As mentioned briefly earlier, the positive class is named ``highlights'' and it represents all the spatio-temporal cuboids starting when a team scores a goal while the negative class (\ie ``standard play'') includes other neutral situations happening during the game. In this work we excluded all the other significant annotated events (fouls, fights, etc.) to reduce the number of classes\footnote{These events are indeed generating different types of excitement, we could not investigate further for lack of annotated data, but this is an argument that we consider worthed of further research.}. In training phase the samples belonging to the two classes have been balanced to avoid dataset bias.

The S-Hock dataset provides a set of synchronized videos of the games including several views of the audience, at different resolution/zoom level, and of the complete game footage. The video acquisition is done from different points of view (frontal and slightly tilted to the side), in this work we used all these views to ensure a more robust model of training that is able to learn features that are more possibly scale and position invariant.
Positive and negative samples are then splitted into training and validation sets with a ratio of 70\%-30\%. Data augmentation procedure has been performed (horizontal flips in the spatial dimension) not only to increase the amount of training data but also to augment the invariance of the network. 

The final optimization is proposed as a classification problem, minimizing the categorical cross-entropy between the two classes. For this procedure we used the \emph{RMSprop} algorithm, a generalization of the resilient backpropagation \emph{rprop} algorithm that extends the ability to use only the sign of the gradient and to adapt the learning rate separately for each weight, to improve the work with minibatches. In our experiments we use minibatches of 64 samples each. A Dropout layer with 50\% probability to disconnect the link is applied before the first two fully connected layers to reduce overfitting.
The procedure iterates over the whole dataset until convergence, usually reached after about 10 epochs.
The whole training procedure takes about 2 hours on a machine equipped with a NVIDIA Tesla K-80 GPU, using Keras/TensorFlow framework. The whole resulting dataset is composed of a total of 32,000 training samples.

\subsection{Quantitative Results}

\begin{figure}[t]
 \centering
 \includegraphics[width=.5\columnwidth]{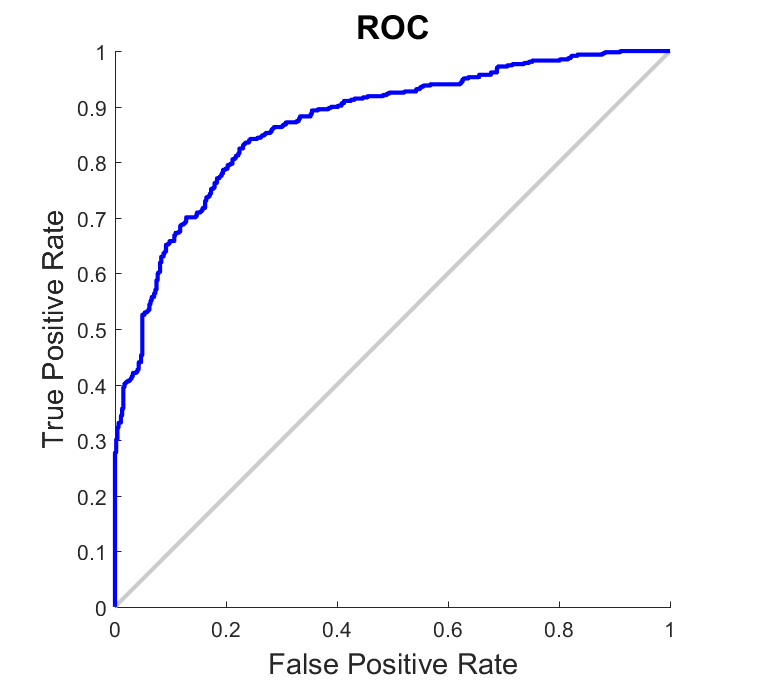}
 \caption{ROC curve}
 \label{fig:res_roc}
\end{figure}

\begin{figure}
	\centering
 	\begin{subfigure}[b]{.49\textwidth}
		\centering
		\includegraphics[width=\columnwidth]{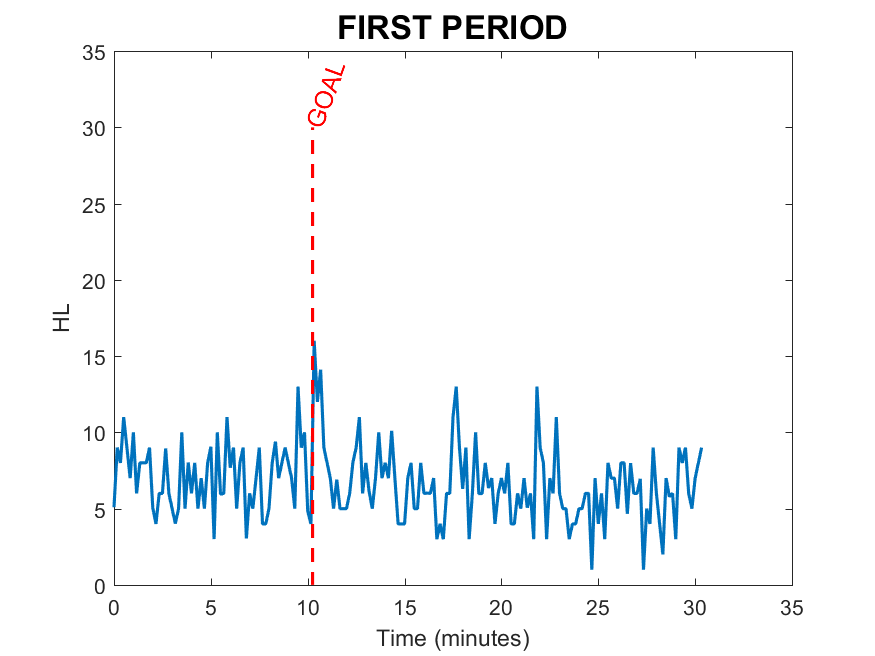}
	\end{subfigure}
	\begin{subfigure}[b]{.49\textwidth}
		\centering
		\includegraphics[width=\columnwidth]{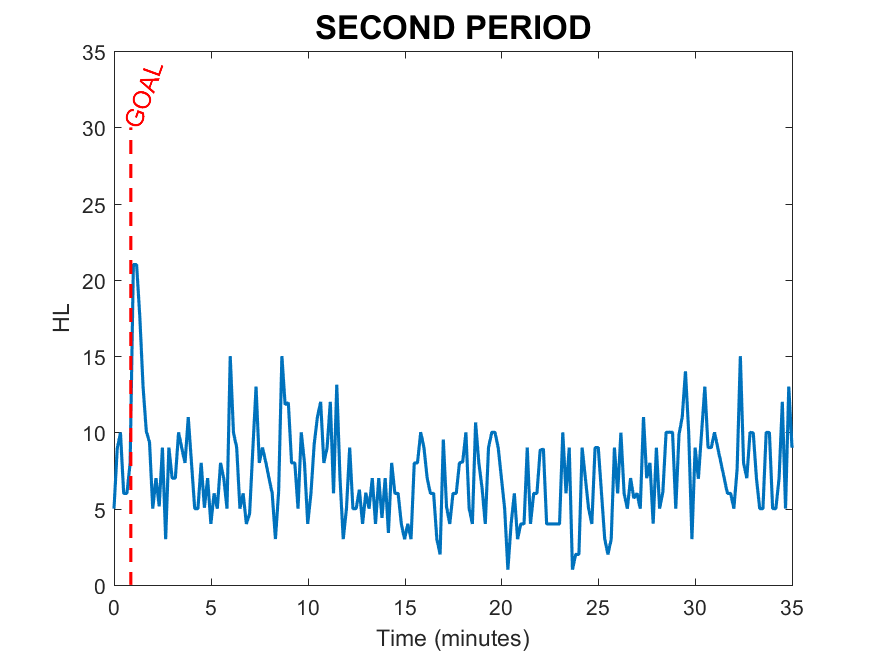}
	\end{subfigure}\\ \vspace{1em}
	\begin{subfigure}[b]{.48\textwidth}
		\centering
		\includegraphics[width=\columnwidth]{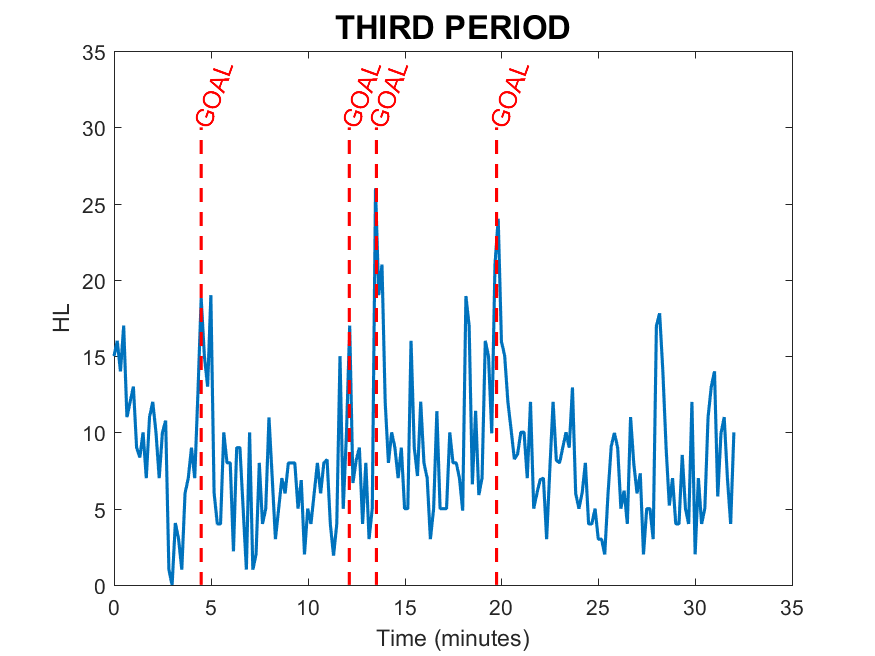}
	\end{subfigure}
	\caption{Summed probabilities of highlights over all the crops in the scene. As visible, peaks in the curve nicely corresponds to a highlight.}
	\label{fig:res_time}
\end{figure}

\begin{figure}
	\centering
	\begin{subfigure}[b]{.47\textwidth}
		\centering
		\includegraphics[width=\columnwidth]{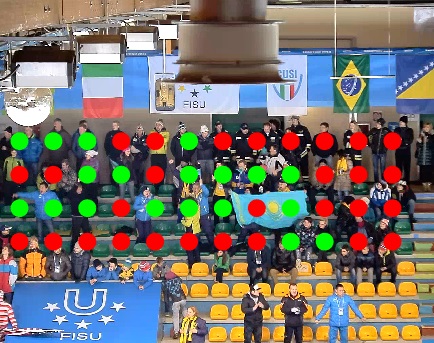}
		\caption{Highlight}
	\end{subfigure}
    \hspace{.04\textwidth}
	\begin{subfigure}[b]{.47\textwidth}
		\centering
		\includegraphics[width=\columnwidth]{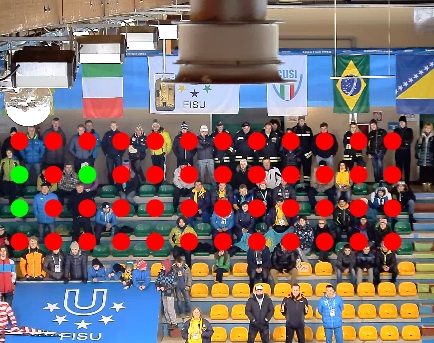}
		\caption{Standard play}
	\end{subfigure}
	\caption{Probability scores given by a subset of the crops (chosen to be non overlapping for visualization purposes); each dot represents a crop which describes part of the scene. Green dots represent crops classified as people reacting to an highlight (\eg cheering) while the red dots represent the crops classified as people with a "standard" behavior.}
	\label{fig:res_dots}
\end{figure}

Here we report a quantitative performance evaluation of the 3D-CNN in detecting positive and negative highlight  samples.
From the second period of the testing game, we randomly selected 3000 positive samples as well as the same number of negative samples and we fed them into the trained network.
In Fig.~\ref{fig:res_roc} the ROC curve is reported. The Area Under the Curve (AUC) is 0.87. Binary classification is performed by assigning the sample to the class corresponding to the higher score; under this conditions the network reaches 78\% of accuracy, 69\% of precision and a recall of 84\%.
Results themselves are quite good considering the difficulty of the task, however, our goal is different, since we are using those results in a more sophisticated framework to infer and rank interesting events during the whole game. Consequently we expect a certain amount of noise in such prediction since in many cases the sample may be partially filled with empty seats (see Fig.~\ref{fig:res_dots} ), producing a wrong or at least biased prediction toward the negative class. However, this problem is minimized with the use of the accumulator approach and due to the fact that the empty-seats location will be very little informative in the whole sequence, while the crowded locations, where most of the spectators are situated, will convey most of the information used for the final decision.

\subsection{Qualitative examples}

We also provide qualitative results to validate our approach.
Fig.~\ref{fig:res_time} shows the HL score, summed over all the cuboids, at every non overlapping 10-second slice during an entire match (3 periods of 20 minutes plus timeouts). 
Goals are clearly identified in the first two periods, while in the third one other events also trigger the audience behavior; in particular, there are two prominent events that don't correspond to goals at 18:45 (which is caused by a player almost scoring) and at 28:15 (which is caused by a foul in front of the goaltender, and the resulting penalty).
We can easily see that there is a correlation between HL score and important events in the game and that goals usually cause the biggest reaction on the spectators.

\section{Conclusions}
\label{sec:concl}

In this paper we propose a method to temporally locate highlights in a sport event by analyzing solely the audience behavior. We propose to use a deep 3D convolutional neural network on cuboid video samples to discriminate between different excitement of the spectators. An spatial accumulator is used to produce a score which is proportional to the probability of having an interesting highlight in that precise time location. This enables the model to identify goals and other salient actions.

Despite being very simple, the model we present provides good preliminary result on a public dataset of hockey games, encouraging further research based on this approach.
In our opinion, the main limit of this model is in the way we take into account the temporal information; indeed we extend a standard CNN to work with 3D data, where the third dimension is time.
A more sophisticated model, such as recurrent neural networks (RNN) and long-short term memory (LSTM), could benefit the final inferential results. As future work we intend to replace the accumulator with such a temporal model, expanding the classification to a multiclass problem in order to detect different events. In order to do so, the dataset has to be enlarged possibly on a different location to make sure the network is learning more general discriminative features.

{\small
	\bibliographystyle{splncs03}
	\bibliography{refs}
	
}

\end{document}